\documentclass[runningheads]{llncs}
\usepackage{graphicx}

\usepackage{tikz}
\usepackage{comment}
\usepackage{amsmath,amssymb} 
\usepackage{color}
\usepackage{kotex}
\usepackage{makecell}
\usepackage{amssymb}
\usepackage{multirow}
\usepackage{wrapfig}
\usepackage{makecell}
\usepackage{ulem}

\usepackage{hyperref}
\hypersetup{colorlinks,allcolors=black}

\usepackage[accsupp]{axessibility}  

\usepackage[width=122mm,left=12mm,paperwidth=146mm,height=193mm,top=12mm,paperheight=217mm]{geometry}

\begin{document}
\pagestyle{headings}
\mainmatter
\def\ECCVSubNumber{5898} 

\title{BigColor: Colorization using \\ a Generative Color Prior for Natural Images} 

\titlerunning{BigColor} 
\authorrunning{G. Kim et al.} 

\author{
        Geonung Kim\inst{1}, \
        Kyoungkook Kang\inst{1}, \
        Seongtae Kim\inst{1}, \
        Hwayoon Lee\inst{1}, \\
        Sehoon Kim\inst{2}, \
        Jonghyun Kim\inst{2}, \  
        Seung-Hwan Baek\inst{1}, \
        Sunghyun Cho\inst{1} \
}
\institute{
        POSTECH\inst{1}  \\
        \email{\{k2woong92,kkang831,seongtae0205,hwayoon2,shwbaek,s.cho\}@postech.ac.kr}\\
        Samsung Electronics\inst{2} \\
        \email{\{sh0264.kim,jh015.kim\}@samsung.com}
\vspace{-5mm}
}

\def\MethodName{BigColor} 

\newcommand{\Eq}[1]  {Eq.\ (#1)}
\newcommand{\Eqs}[1] {Eqs.\ (#1)}
\newcommand{\Fig}[1] {Fig.\ #1}
\newcommand{\Figs}[1]{Figs.\ #1}
\newcommand{\Tbl}[1]  {Tab.\ #1}
\newcommand{\Tbls}[1] {Tabs.\ #1}
\newcommand{\Sec}[1] {Sec.\ #1}
\newcommand{\SSec}[1] {Sec.\ #1}
\newcommand{\Secs}[1] {Secs.\ #1}
\newcommand{\Alg}[1] {Alg.\ #1}
\newcommand{\etal}   {{\textit{et al.}}}

\renewcommand{\topfraction}{0.95}
\setcounter{bottomnumber}{1}
\renewcommand{\bottomfraction}{0.95}
\setcounter{totalnumber}{3}
\renewcommand{\textfraction}{0.05}
\renewcommand{\floatpagefraction}{0.95}
\setcounter{dbltopnumber}{2}
\renewcommand{\dbltopfraction}{0.95}
\renewcommand{\dblfloatpagefraction}{0.95}

\newcommand{\Net}[1]{#1}
\newcommand{\Loss}[1]{$\mathcal{L}_{#1}$}
\newcommand{\cm}{\checkmark}
\newcommand\oast{\stackMath\mathbin{\stackinset{c}{0ex}{c}{0ex}{\ast}{\bigcirc}}}

\renewcommand{\paragraph}[1]{{\vspace{2pt}\noindent\textbf{#1}}}

\maketitle

\begin{abstract}
For realistic and vivid colorization, generative priors have recently been exploited.
However, such generative priors often fail for in-the-wild complex images due to their limited representation space.
In this paper, we propose \MethodName, a novel colorization approach that provides vivid colorization for diverse in-the-wild images with complex structures. While previous generative priors are trained to synthesize both image structures and colors, we learn a generative color prior to focus on color synthesis given the spatial structure of an image. In this way, we reduce the burden of synthesizing image structures from the generative prior and expand its representation space to cover diverse images. 
To this end, we propose a BigGAN-inspired encoder-generator network that uses a spatial feature map instead of a spatially-flattened BigGAN latent code, resulting in an enlarged representation space. 
\MethodName\ enables robust colorization for diverse inputs in a single forward pass, supports arbitrary input resolutions, and provides multi-modal colorization results. 
We demonstrate that \MethodName\ significantly outperforms existing methods especially on in-the-wild images with complex structures.

\if{0}
For realistic and vivid colorization, generative priors have recently been exploited.
However, such generative priors often fail for in-the-wild complex images due to their limited representation space.
In this paper, we propose \MethodName, a novel colorization approach that provides vivid colorization of diverse in-the-wild images with complex structures. While previous generative priors are trained to synthesize both image structures and colors, we learn a generative color prior to focus only on color synthesis given the spatial structure of an image. In this way, we reduce the burden of synthesizing image structures from the generative prior and expand its representation space to cover diverse images. Based on the generative color prior, we also propose an encoder-generator colorization network that can colorize an image in a single forward pass, and a multi-modal color synthesis scheme. Our experiments demonstrate that \MethodName~significantly outperforms existing methods especially on in-the-wild images with complex structures.
\fi

\end{abstract}

\section{Introduction}
\label{sec:introduction}

Image colorization aims to hallucinate the chromatic dimension of a grayscale image and has been studied for decades in computer vision and graphics.
Its application includes not only modernizing classic black-and-white films but also providing artistic control over grayscale imagery with diverse color distributions
\cite{Rephotography,CIC,charpiat2008automatic,TowardVivid,ColTran}.

\begin{figure}
\centering
\includegraphics[width=12cm]{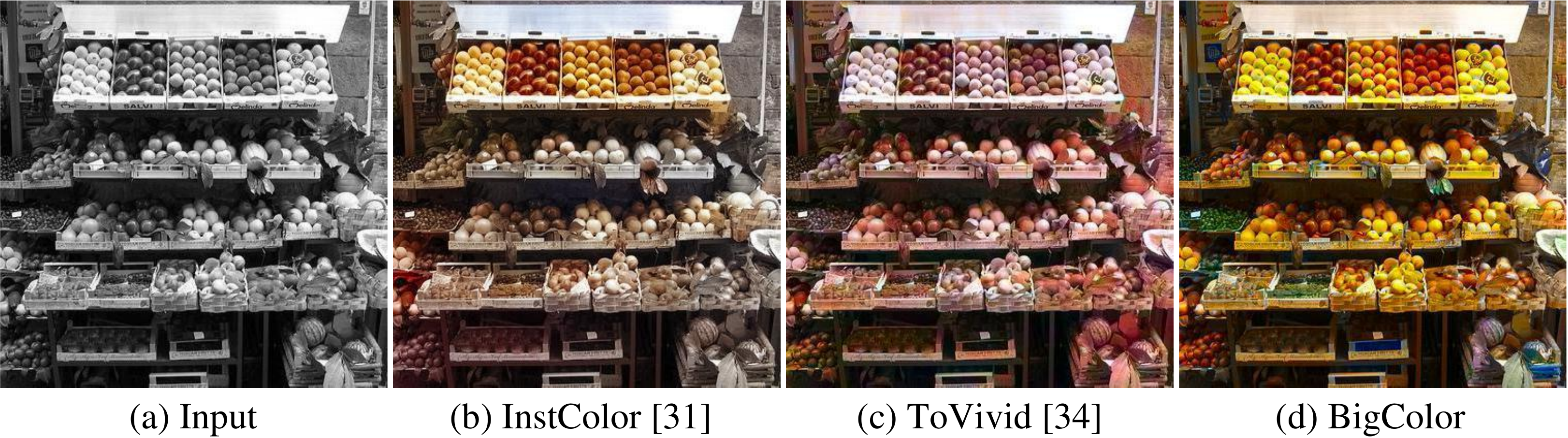}
\caption{
We achieve robust colorization for in-the-wild images using a generative color prior.
(a) For an input image with complex spatial structures, existing colorization methods suffer from (b) desaturated color and (c) unnatural color distribution. (d) In contrast, \MethodName\ synthesizes natural colors consistent with the input structure using a learned generative color prior.
}
\vspace{-5mm}
\label{fig:teaser}
\end{figure}

Early works propagate user-annotated color strokes based on pixel affinity~\cite{levin2004colorization,huang2005adaptive,yatziv2006fast,qu2006manga,xu2009efficient} or find similar regions in reference images to mimic the reference color distributions~\cite{charpiat2008automatic,chia2011semantic,gupta2012image}.
With the advent of deep learning, data-driven colorization approaches have rapidly advanced by adopting neural networks to learn a mapping from grayscale images to trichromatic images.
This trend was sparked by using a convolutional neural network (CNN) and a regression loss such as mean-squared error (MSE)~\cite{CIC,ChromaGAN,InstColor,Deoldify}, which unfortunately suffers from desaturated colors as shown in \Fig{\ref{fig:teaser}} (b), as the MSE loss encourages to find an average of plausible color images corresponding to an input image.


To synthesize vivid colors, high-quality representations learned in pretrained generative adversarial network (GAN) models have recently been exploited as generative priors for image colorization~\cite{TowardVivid,GPEN,GFP-GAN,DGP,mGANPrior}.
Adopting GAN inversion, these methods invert an input grayscale image to a latent code of a pretrained GAN model by minimizing the structural discrepancy between the input gray-scale image and the generated color image from the latent code. 
While GAN inversion allows us to utilize the learned generative prior of natural images, it also inherits a notable problem of existing GAN models: limited representation space. 
Thus, existing colorization methods using generative priors fail to handle in-the-wild images with complex structures and semantics, resulting in desaturated and unnatural colors as shown in \Fig{\ref{fig:teaser}}(c).

In this paper, we propose \MethodName, a novel image colorization method that synthesizes vivid and natural colors for in-the-wild images with complex structures.
For vivid colorization, we adopt the GAN-inversion approach by using a pretrained BigGAN~\cite{BigGAN}, which is a state-of-the-art class-conditional generative model.
As directly using the BigGAN model hampers colorization performance for in-the-wild images due to its limited representation space, 
we offload the burden of the BigGAN model that was responsible for synthesizing both structures and colors to focus on color synthesis. 
This offloading strategy allows us to learn a generative color prior that can cover in-the-wild images with complex structures.
Specifically, we learn a generative color prior with an encoder-generator neural network. 
Unlike conventional GAN-inversion colorization methods, our encoder extracts a spatial feature map describing the structure of an input image better than using a spatially-flattened latent code in BigGAN.
As a spatial feature map has a higher spatial resolution than an original BigGAN latent code, the representation space of the entire network can be enlarged, i.e., we can map features to a wider range of natural images.
We then design our generator to directly exploit the spatial feature by using the fine-scale network layers adopted from the multi-scale BigGAN generator. 
We jointly train the encoder and generator networks to encourage the network to focus on color synthesis by making use of the spatial feature. %
As our network is fully convolutional and departs from using a fixed-size flattened latent code of BigGAN, \MethodName\ can process images with arbitrary sizes which were not feasible for conventional GAN-inversion colorization methods that use the original latent codes of GANs~\cite{DGP,TowardVivid,mGANPrior,GPEN,GFP-GAN}.
Also, \MethodName\ allows us to synthesize multi-modal colorization results by using different condition vectors for the network.
We assess \MethodName\ with extensive experiments including a user study and demonstrate that \MethodName\ outperforms previous methods across all tested scenarios in particular for in-the-wild images.

\if{0}
Image colorization hallucinates the chromatic dimension of a grayscale image and has been studied for decades in computer vision and graphics.
Its application includes not only modernizing classic black-and-white films but also providing artistic control over grayscale imagery with diverse color distributions
\cite{Rephotography,CIC,charpiat2008automatic,TowardVivid,ColTran}.

Early works propagate user-annotated color strokes based on pixel affinity~\cite{levin2004colorization,huang2005adaptive,yatziv2006fast,qu2006manga,xu2009efficient} or find similar regions in reference images to mimic the reference color distributions~\cite{charpiat2008automatic,chia2011semantic,gupta2012image}.
With the advent of deep learning, data-driven colorization approaches have rapidly advanced by adopting neural networks to learn a mapping from grayscale images to trichromatic images.
This trend was sparked by using a convolutional neural network (CNN) and a regression loss such as MSE~\cite{CIC,ChromaGAN,InstColor,Deoldify}, which unfortunately suffer from unnatural desaturated colors, as the MSE loss encourages to find an average of plausible color images corresponding to an input image as shown in \Fig{\ref{fig:teaser}} (b).

To recover vivid colors,
the high-quality representation learned in pretrained generative adversarial network (GAN) models has recently been exploited as generative priors for image colorization~\cite{TowardVivid,GPEN,GFP-GAN,DGP,mGANPrior}.
Adopting GAN inversion, these methods invert an input grayscale image to a latent code in the latent space of a pretrained GAN model that is mapped to a most similar color image to the input grayscale image.
While GAN inversion allows us to utilize the learned generative prior of structures and colors of natural images, the limited representation space of pretrained GAN models poses a notable challenge in its inversion process especially for in-the-wild images with complex structures and semantics.
Thus, existing colorization methods using generative priors fail to handle such challenging images suffering from desaturated colors, color bleeding\kkw{color bleeding을 지우고 unnatural color만 남기는 것이 어떨까요.} and semantically unnatural colors as shown in \Fig{\ref{fig:teaser}}(c).

In this paper, we propose \MethodName, a novel image colorization method using a BigGAN~\cite{BigGAN}-based \emph{generative color prior} that provides vivid and semantically matched colorization over in-the-wild images with complex structures.
For vivid colorization of natural images of diverse classes,
our approach adopts a pretrained BigGAN generator, which is a state-of-the-art class-conditional generative model.
Directly using a BigGAN generator as a generative prior, however, may result in poor colorization performance due to its limited representation space.
To overcome this and to cover diverse in-the-wild images,
we learn a generative color prior from a pretrained BigGAN generator by refining it to specialize in synthesizing color instead of putting its efforts on generating both colors and structures.
In this way, we can reduce the burden of learning the joint distribution of structures and colors of natural images from the generative prior and expand its representation space to cover diverse images.

We build our specialized generative color prior as follows.
First, we modify the network architecture of the BigGAN generator by taking only fine scales from its multi-scale architecture.
The modified architecture takes a spatial feature map that encodes the spatial structure of an image instead of a latent code in the original latent space.
Second, we build an encoder network to extract a spatial feature map from an input image, which serves as an input to the modified generator.
We then jointly train the encoder and generator networks to make the network to more focus on synthesizing colors for the input spatial feature map instead of synthesizing image details or structures.
As the spatial feature map of a fine scale has a higher dimension than the original latent code of BigGAN, we can effectively enlarge the representation space, i.e., we can map latent codes to a wider range of natural images.
Moreover, due to the rich information on the image structure contained in the spatial feature map and the joint training of the encoder-generator networks, the generator can solely focus on color synthesis without the burden of synthesizing image structures, which eventually leads to high-quality colorization of diverse images.

Based on the generative color prior,
we design our framework as a encoder-generator architecture where the generator acts as a decoder.
Our framework is fully convolutional and able to handle images of arbitrary sizes.
we also present a class-conditional encoder architecture for effective estimation of the spatial feature map from images of multiple classes.

Besides, we propose a multi-modal colorization scheme to enhance the diversity of colorization and to provide users to control over the colorization result.
We also present a color augmentation-based training scheme, which is simple but effectively improves the color vividness.
We validate the proposed method through extensive evaluations including a user study and demonstrate that our method outperforms previous methods across all tested scenarios in particular for in-the-wild images.

\fi

\if{0}
Image colorization is the problem estimating colored image from grayscale image.
Previous deep learning based colorization methods \cite{zhang2016colorful,vitoria2020chromagan,su2020instance} train neural networks model using loss function comparing the pixel-wise differences between ground truth colored image and inferred one.
Although they show the impressive colorization results, Training without taking into account any semantic information results in inconsistent and unnatural colorization. Additionally, colorization is inherently ill-posed problem so the color ambiguity results in desaturated and grayish image.

Recently, in order to generate more accurate and vivid colored image, various methods propose to utilize the realistic image generation capability of Generative Adversarial Nets\cite{goodfellow2014generative} as a prior knowledge for the image colorization problem \cite{yang2021gan,wang2021towards,wu2021towards,gu2020image,pan2021exploiting}.
GANs are trained to create a high quality realistic image in a structure in which a generator seeking to create a realistic image from a latent vector $z$ and a discriminator that distinguishes the generated image from the real image compete with each other.
The generative models know plenty semantic information and color distribution of the real images so that it is useful to use the GANs as prior knowledge.
The generative prior is mainly used through GAN inversion, which is to find a latent code of GAN that can generate a corresponding image for a given input image.
In the case of image colorization, a given grayscale image is projected into a latent code of GANs that produces the same structure, and the generator decode it for generating colored image.

StyleGAN\cite{karras2019style,karras2020analyzing} and BigGAN\cite{brock2018large} are the most widely used GAN models for the generative prior.
With StyleGAN, \cite{yang2021gan,wang2021towards,luo2021time} succeeded in colorizing grayscale images with high quality.
However, Their colorization capability is limited to a single class of images because StyleGAN is generally trained using only one class such as human face or car.
On the other hand. BigGAN is trained using ImageNet1K dataset\cite{russakovsky2015imagenet} so that the model have generative capability for general image distribution. With this capability, \cite{wu2021towards,pan2021exploiting} propose a colorizing method for general grayscale images using BigGAN.
However, the latent space $Z$ of BigGAN has limited representation range so that the inversion results cannot reconstruct the original image structure.
In order to overcome the poor limitation \cite{wu2021towards} use feature alignment process and \cite{pan2021exploiting} jointly optimize the latent code $z$ and parameters of Generator.
Despite this efforts, poor color results remain a problem.

Recently, in order to address limited representation problem of latent space, a method using the intermediate feature of GANs as an inversion latent space has been proposed \cite{kang2021gan}.
The GAN's intermediate feature has much greater representational power than the existing GAN's latent code $z$, which greatly increases the range of images that can be treated in the GAN inversion.

In this paper, we propose a grayscale image colorization method using the feature space GAN inversion of BigGAN.
An encoder for projecting an input grayscale image into an intermediate feature is designed and combined with the intermediate layer of BigGAN generator. we show that \MethodName outperforms privious state-of-the-art methods for various metric.
Next, through various experiments, it is shown that the results of this paper overcome the limitation of damaging the structure of the image when it is thought of as the existing latent space Z, and greatly increase the image colorization performance for general images. 

Our main contributions can be summarized as follows.
\begin{itemize}
    \vspace{-0.3cm}
    \item We propose \MethodName, a novel GAN inversion approach using intermediate feature as inversion target for grayscale image colorization.
    \item We introduce class conditional GAN inversion encoder architecture.
    \item Qualitative and qualitative results show that our method achieves state-of-the-art performance on image colorization outperforming prior methods.
\end{itemize}
\fi

\section{Related Work}

\paragraph{Optimization-based Colorization}
Early colorization methods utilize color annotations from users and propagate them to neighbor pixels based on pixel affinity by solving constrained optimization problems~\cite{levin2004colorization,huang2005adaptive,yatziv2006fast,qu2006manga,xu2009efficient}.
Data-driven colorization methods find reference color images with similar semantics to an input grayscale image and use the reference color distributions via optimization~\cite{liu2008intrinsic,charpiat2008automatic,chia2011semantic,gupta2012image}.
Unfortunately, the optimization-based approaches demand dense user annotations or accurate reference matching, failing to provide robust and automatic colorization. 

\paragraph{Colorization with Regression Networks}
Learning a mapping function from a grayscale image to a color image has been extensively studied with the advent of neural networks.
Regression-based neural networks minimize average reconstruction error, resulting in desaturated colors~\cite{DeepColor,LargeAutomaticColor,LearningAumotaticColor,LetThereBeColor}.
Vivid color synthesis then became one of the core challenges in network-based image colorization methods.
Notable examples in this line of research include optimizing over a quantized color space~\cite{CIC}, detection-guided colorization~\cite{InstColor}, adversarial training~\cite{ChromaGAN,Deoldify}, and global reasoning using a transformer~\cite{ColTran}.
While significant progress has been made, it is still challenging to synthesize vivid and natural colors for in-the-wild grayscale images with complex structures.

\paragraph{Colorization with Generative Prior}
GANs have recently achieved remarkable success in learning low-dimensional latent representations of natural color images, enabling synthesizing high-fidelity natural images~\cite{StyleGAN,StyleGAN2,BigGAN}. 
This success has led to using the learned generative prior for image restoration such as deblurring~\cite{GFP-GAN,GPEN}, super-resolution~\cite{GLEAN,PULSE,DGP}, denoising~\cite{GFP-GAN,GPEN}, and colorization~\cite{GFP-GAN,GPEN,DGP,TowardVivid,mGANPrior}.
Most previous approaches are limited to handling a single class of images, such as human faces using StyleGAN~\cite{StyleGAN,StyleGAN2}, due to the limited representation space of modern GAN models. 

Recently, a few attempts~\cite{DGP,TowardVivid} have been made to colorize natural images of multiple classes using a pretrained BigGAN generator~\cite{BigGAN}.
Specifically, deep generative prior (DGP)~\cite{DGP} jointly optimizes the BigGAN latent code and the pretrained BigGAN generator to synthesize a color image via GAN inversion.
The representation space of the DGP is still not enough to cover complex images because of the difficulty in synthesizing both structures and colors from the generator. 
Wu et al.~\cite{TowardVivid} attempted to bypass the structural mismatch between a GAN-inverted color image and an input grayscale image by warping the synthesized color features into the input grayscale. 
Nonetheless, considerable mismatches between a GAN-inverted and an input image cannot be fully resolved, and thus produce colorization artifacts.
In contrast to the previous methods, \MethodName\ effectively enlarges the representation space by using an encoder-generator architecture that uses spatial features.
This allows us to handle diverse images with complex structures.

\if{0}
\section{Related Work}

\paragraph{Optimization-based Colorization.}
Early colorization methods utilize user annotations of color and propagate them to neighbour pixels based on pixel affinity by solving constrained optimization problems~\cite{levin2004colorization,huang2005adaptive,yatziv2006fast,qu2006manga,xu2009efficient}.
Data-driven methods find reference color images with similar semantics to an input grayscale image and utilize the color distribution of the reference images for image colorization~\cite{liu2008intrinsic,charpiat2008automatic,chia2011semantic,gupta2012image}.
Unfortunately, the optimization-based approaches demand dense user annotations or reference matching due to their lack of prior knowledge on the general color distribution of natural images.

\paragraph{Colorization with Regression Networks.}
Learning a mapping function from a grayscale image to a color image has been extensively studied with the advent of neural networks.
Regression-based neural networks minimize the average reconstruction error, resulting in desaturated colors~\cite{DeepColor,LargeAutomaticColor,LearningAumotaticColor,LetThereBeColor}.
Vivid color synthesis then becomes one of the core challenges in network-based image colorization methods.
Notable examples in this line of research include optimizing over a quantized color space~\cite{CIC}, detection-guided colorization~\cite{InstColor}, adversarial training~\cite{ChromaGAN,Deoldify}, and global reasoning using a transformer~\cite{ColTran}.
While notable progress has been made, it is still challenging to synthesize plausible natural colors for in-the-wild grayscale images with complex structures.

\paragraph{Colorization with Generative Prior.}
GANs have recently achieved remarkable success in learning low-dimensional latent representations of natural color images, allowing of synthesizing high-fidelity natural images~\cite{StyleGAN,StyleGAN2,BigGAN}. 
This success has led to utilizing the learned generative prior for image restoration tasks including deblurring~\cite{GFP-GAN,GPEN}, super-resolution~\cite{GLEAN,PULSE,DGP}, denoising~\cite{GFP-GAN,GPEN}, and colorization~\cite{GFP-GAN,GPEN,DGP,TowardVivid,mGANPrior}.
However, most of the previous approaches are limited to a single class of images, such as human faces using StyleGAN~\cite{StyleGAN,StyleGAN2}, due to the limited representation space of modern GAN models. 

Recently, a few attempts~\cite{DGP,TowardVivid} have been made for the restoration of multiple classes of natural images using a pretrained BigGAN generator~\cite{BigGAN}.
However, these methods also struggle with the limited representation space of BigGAN. 
Specifically, the deep generative prior~\cite{DGP} performs optimization of the latent code and the network parameters of a pretrained BigGAN generator for each input image to perform image restoration including colorization.
Wu et al.~\cite{TowardVivid} proposed a BigGAN-based colorization approach that finds a reference image similar to an input image using BigGAN inversion.
Due to the limited representation space of BigGAN, an estimated reference image is often considerably different from an input image.
To resolve the gap between the reference and input images, their approach introduces a warping layer based on spatial feature correlation between a reference and an input image.
Nonetheless, considerable mismatches between a reference and an input image cannot be resolved perfectly, and may produce colorization artifacts.
On the other hand, our approach effectively enlarges the representation space, thus can handle diverse images with complex structures.

\fi

\if{0}
\subsubsection{Classic Colorization Methods}
Previously, colorization methods mostly require user guide or reference image and use these additional materials as prior knowledge for reasonable color generation.
In order to find missing color, the propagation based methods use color strokes\cite{levin2004colorization,huang2005adaptive,yatziv2006fast,qu2006manga,xu2009efficient} or color points\cite{zhang2017real} from an user.
The local color information extends to semantically related region, which fill with intended color using optimization process.
On the other hand, the reference-based colorization methods\cite{liu2008intrinsic,charpiat2008automatic,chia2011semantic,gupta2012image,he2018deep,chia2011semantic} use the image itself as color prior.
they find semantically corresponding region between colored reference image and black-and-white target image and colorize the gray image based on the estimated reference colors. 
Although they succeed in generating desirable and correct colors, the additional inputs from user makes automation impossible.

\subsubsection{Automatic Colorization Methods}
Recently, various approaches for image colorization have been proposed and are divided into the presence or absence of a generative prior.
The methods without generative prior use not only simple date-term loss but additional techniques to overcome the limitations such as desaturated or semantically incorrect color results. 
CIC\cite{CIC} proposes quantized color space and color class rebalancing to generate vivid colors.
InstColor\cite{InstColor} leverages an off-the-shelf object detector and separately handles objects and background for reasonable color prediction.
ChromaGAN\cite{ChromaGAN} and Deoldify\cite{Deoldify} train colorization network with adversarial loss using discriminator for more realistic and semantic-aware colors.
ColTran\cite{ColTran} firstly estimates colors at low resolution and then upsamples both color and spatial resolution using transformer based architecture.
Although they show the impressive colorization results, inconsistent and desaturated color results remain unsolved.

\sunghyun{다음 내용 여기에 잘 녹여 넣기}
As a remedy for the desaturation problem, adopting the adversarial loss~\cite{GAN} reduces the domain gap between the resulting color images and the natural color images~\cite{ChromaGAN,Deoldify}.

\subsubsection{Colorization with Generative Prior}
The colorization methods with generative prior is mainly used in the form of GAN inversion\cite{Image2StyleGAN,MulticodeGAN,IndomainGAN} and succeed in producing high quality color results with some limitations.
GPEN\cite{GPEN} and GFP-GAN\cite{GFP-GAN} show face image colorization based on GAN inversion with fine-tuning using pretrained StyleGAN\cite{StyleGAN, StyleGAN2}.
Although they show the colorization results indistinguishable from real image, it is limited to a single category.
DGP\cite{DGP} propose an optimization method preserving generative prior using BigGAN\cite{BigGAN} and show various image colorization results, but the colorization frequently fails due to the hardness of inversion with BigGAN which has insufficient representational capacity.
ToVivid\cite{TowardVivid} introduce feature warping and fusion in order to overcome the difficulty of inversion, but they also fail when the semantic information does not match between inversion target and result image.

Recently, BDInvert\cite{BDInvert} propose alternative latent space including intermediate feature map space to extend representational capacity of GANs.
They show the effectiveness of the feature space inversion to representational expansion of GANs due to the large dimension and spatial information.
Inspired by BDInvert, we propose GAN inversion-based automatic colorization method using feature space as inversion target in order to alliviate previously unsolved bottleneck, difficutly of GAN inversion.
\fi
\section{Colorization using a Generative Color Prior}

\begin{figure}[t]
\centering
\includegraphics[width=11.5cm]{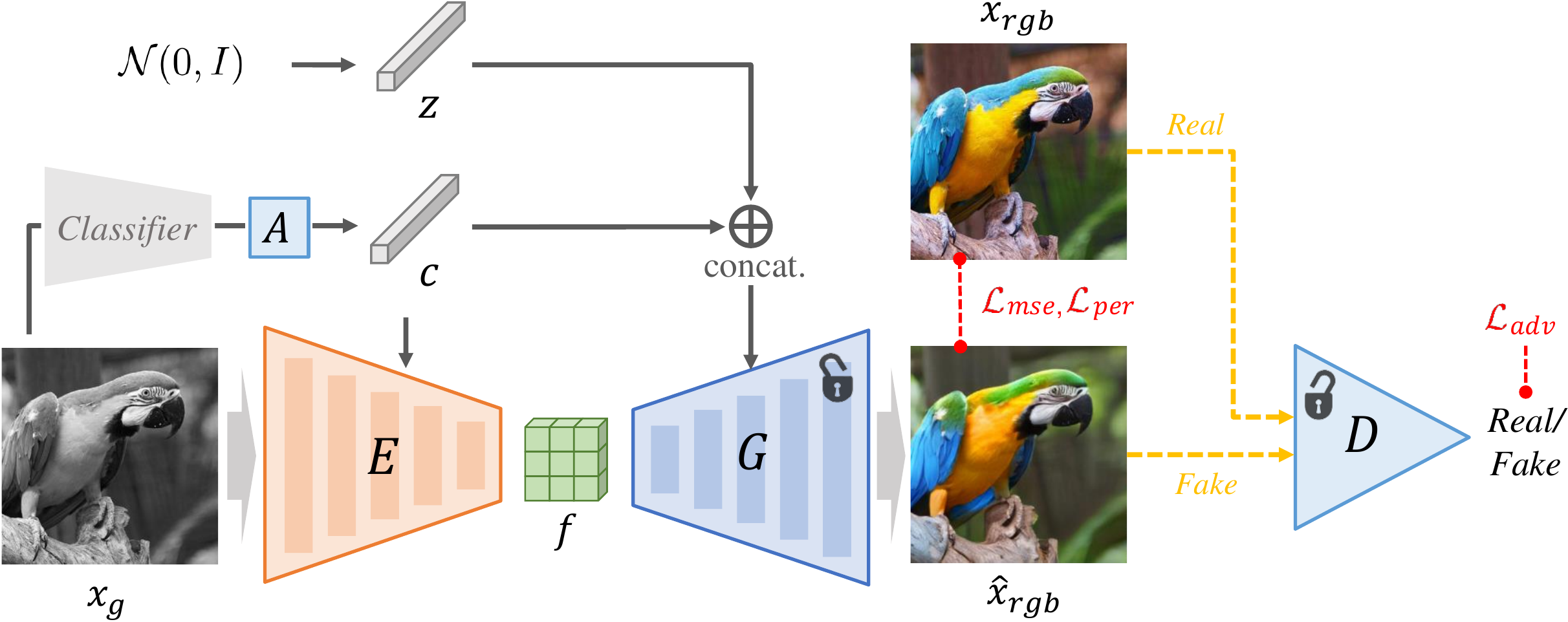}
\caption{
We extract the spatial feature $f$ of the input image $x_g$ using a class-conditioned convolutional encoder $E$. The generator $G$, which is initialized with the fine levels of the pretrained BigGAN~\cite{BigGAN}, takes the spatial feature $f$ as inputs and synthesizes a colorized image $\hat{x}_{rgb}$ conditioned on the control parameters of the class code $c$ and the random sample $z$. $A$ is a class embedding layer that transforms a one-hot class vector to a class code $c$. Jointly training the encoder-generator model with a pretrained BigGAN discriminator $D$ enables us to learn the generative color prior with an enlarged representation space. 
}
\vspace{-3mm}
\label{fig:framework}
\end{figure}

In this section, we describe the framework of \MethodName~and our strategy to learn a generative color prior.
\MethodName~has an encoder-generator network architecture, where the encoder $E$ estimates a spatial feature map $f$ from an input grayscale image $x_g$, and the generator $G$ synthesizes a color image $\hat{x}_{rgb}$ from the feature $f$. 
Note that different from conventional GAN-based colorization methods, we do not rely on the spatially-flattened latent code of BigGAN, but instead use a spatial feature map $f$ that has a larger dimension. 
In order to exploit the effectiveness of the BigGAN architecture for image synthesis~\cite{BigGAN}, we design the encoder $E$ and the generator $G$ by using the fine-scale layers of the BigGAN generator.
Also, we use two control variables for conditioning the encoder and the generator: the class code $c$ and the random code $z$ sampled from a normal distribution. 
The class code $c$ enables class-specific feature extraction for effective colorization and the random code $z$ accounts for the multi-modal nature of image colorization.

In the spirit of adversarial learning, we also adopt a pretrained BigGAN discriminator $D$.
We jointly train the encoder $E$, the generator $G$, and the discriminator $D$, resulting in an enlarged representation space where the generator $G$ takes the responsibility of synthesizing color on top of the spatial feature $f$ extracted from the encoder $E$.
See \Fig{\ref{fig:framework}} for an overview of \MethodName. 
In the following, we describe each component of \MethodName~and the training scheme in detail.

\subsection{Encoder}

\begin{figure}[t]
\centering
    \includegraphics[width=12cm]{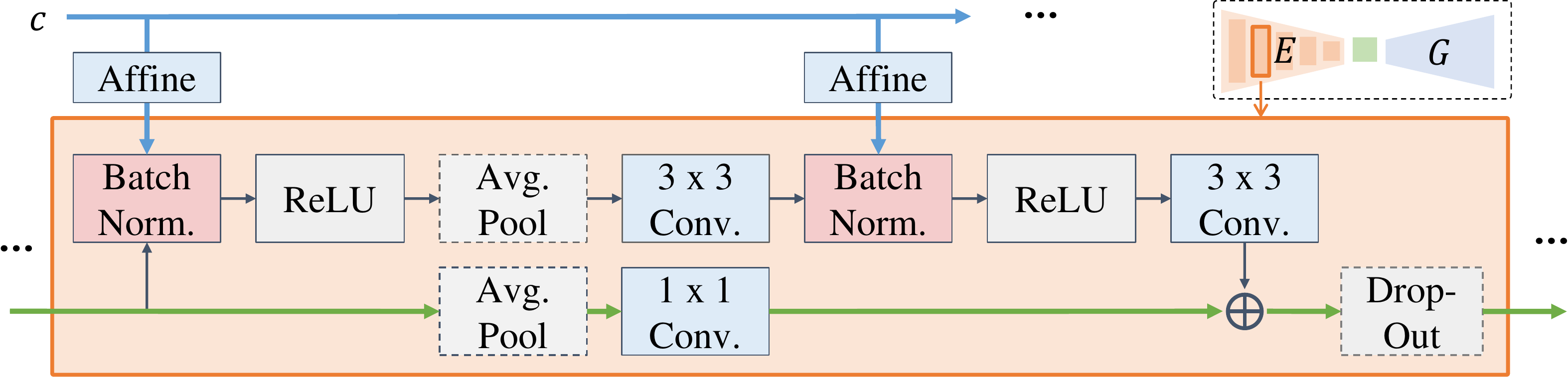}
  \caption{\label{fig:encoder}
    We design our encoder $E$ by inverting the fine layers of the BigGAN generator~\cite{BigGAN}, consisting of the five encoder blocks shown in the top right. Each encoder block denoted as an orange box extracts the spatial features conditioned on the class code $c$. The first and the last encoder blocks do not have average pooling and dropout respectively. 
    }
\vspace{-3mm}
\end{figure}

Our encoder takes an input grayscale image $x_g$ and estimates a spatial feature map $f$, which is fed to the generator.
For an input image size of $256\times256$, our spatial feature $f$ has the spatial resolution of $16\times16$ with the channel size of 768.
To successfully extract the spatial feature $f$, we design our encoder inspired by an inversion of the BigGAN generator as shown in \Fig{\ref{fig:encoder}}. 
The encoder consists of five blocks, where all the blocks except for the first have average pooling layers to reduce the spatial size of an input feature.
We also adopt dropout layers except for the last block for better generalization on test-case inputs.

To extract class-specific spatial structures, we inject the class information of an input image into the encoder.
Specifically, we obtain the scale and bias parameters of the batch-normalization layers through an affine transformation of the BigGAN class code $c \in \mathbb{R}^{128\times1}$~\cite{BigGAN}.
We adopt the BigGAN's class embedding layer ($A$ in \Fig{\ref{fig:framework}}) to obtain the class code $c$ from a class vector in the form of the one-hot vector representation.
The class vector can be either provided by the user or estimated using an off-the-shelf classifier.
In our experiments, we use a 1,000-dimensional vector for the class vector representing ImageNet-1K classes.
More details on the architecture can be found in the Supplemental Document.
In summary, our encoder $E$ extracts the class-specific spatial feature map $f$ that contains the structure information of an input image $x_g$ as 
\begin{equation}
f=E(x_g;c).
\end{equation}

\subsection{Generator}
\begin{figure}[t]
\centering
\includegraphics[width=12cm]{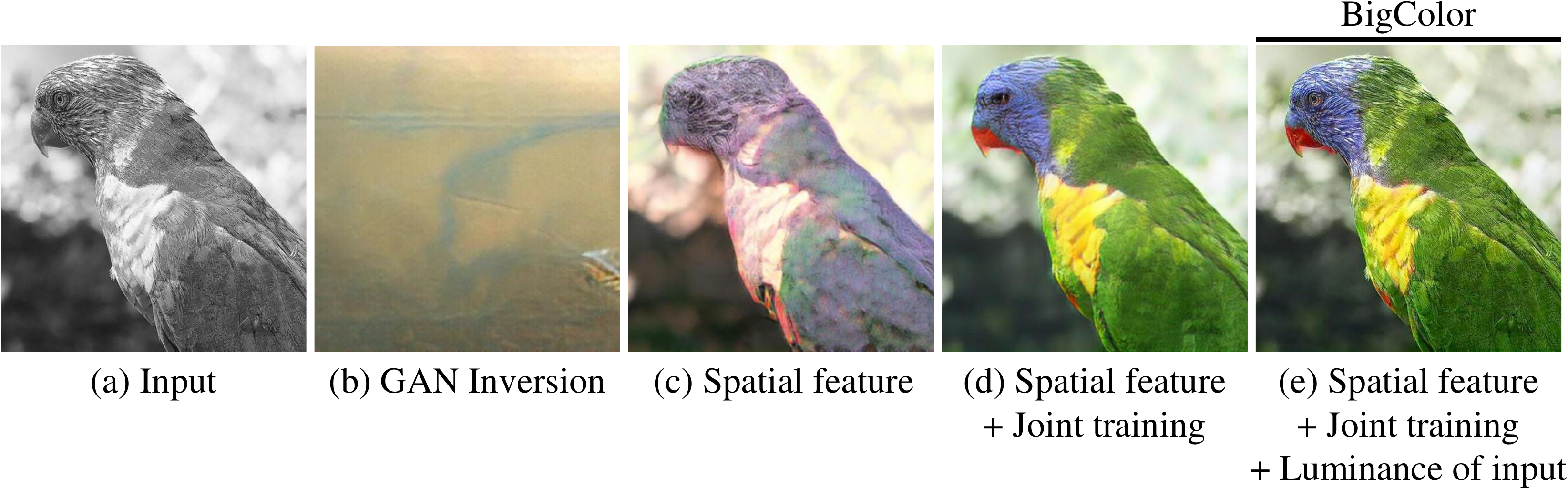}
\caption{
(a)\&(b) Colorization with conventional GAN inversion often fail to invert in-the-wild images. 
(c) We exploit the spatial feature of the input and the fine-scale layers of the pretrained BigGAN generator, effectively enlarging the representation space. 
(d) Jointly optimizing the encoder-generator module improves the representation coverage and provides vivid and natural colorization. 
(e) We boost high-frequency details by replacing the luminance of synthesized image with the input luminance. 
\vspace{-3mm}
}
\label{fig:ablation_fdim_teaser}
\end{figure}

Our generator $G$ synthesizes colors given the spatial feature $f$ of the input gray-scale image $x_g$.
Analogously to the encoder design, we design and initialize our generator $G$ using the fine-scale layers of the pretrained BigGAN generator, specifically from the third to the last layers.
The generator $G$ uses two condition variables of the class vector $c$ and the random vector $z$ sampled from a normal distribution.
We concatenate the class vector $c$ and the random vector $z$ as an input to the generator $G$ as in the original BigGAN architecture~\cite{BigGAN}.
Our generator $G$ synthesizes a color image $\hat{x}_{rgb}$ conditioned on the class and the random codes as 
\begin{equation}
\hat{x}_{rgb}=G(f;c,z).
\end{equation}
We note that unlike the original BigGAN generator that uses a spatially-flattend latent code, our generator $G$ takes the \textit{spatial} feature $f$ as input.
 To restore high-frequency spatial details, we replace the luminance of the synthesized color image $\hat{x}_{rgb}$ with the luminance of the input grayscale image $x_g$ in the CIELAB color space~\cite{CIC,ChromaGAN,InstColor}. See Fig.~\ref{fig:ablation_fdim_teaser}(e).

\paragraph{Generative Color Prior}
We learn the generative color prior for colorizing in-the-wild images with complex structures using our generator $G$. To this end, we exploit our specific network architecture and training scheme.
For the architecture, our generator $G$ takes the fine-scale spatial feature map $f$ as an input of which resolution is $16\times16\times768$ when the grayscale image has $256\times256$ resolution.
The dimension of the feature $f$ is higher than that of the original BigGAN latent code of which resolution is $119\times1$. 
Thus, we can effectively enlarge the representation space of our generator $G$ compared to the conventional GAN-inversion colorization methods by utilizing the structural information provided in the large-dimensional feature $f$.
Compare the colorization results of Fig.~\ref{fig:ablation_fdim_teaser}(b)\&(c).
We note that a similar finding was used in BDInvert~\cite{BDInvert}, a recent transform-robust GAN-inversion method using a spatial feature for StyleGAN~\cite{StyleGAN,StyleGAN2}. 

In terms of training strategy, we initialize the generator $G$ and the discriminator $D$ with the corresponding layers of the ImageNet-pretrained BigGAN model. As such, we can leverage the learned structure-color distribution of natural images of the pretrained BigGAN. 
However, our generator $G$ at the initial point is still not fully focusing on synthesizing colors as it was originally trained to synthesize both structure and color. 
We unlock the full potential of our network by jointly optimizing the encoder $E$, the generator $G$, and the discriminator $D$.
The joint training allows the generator $G$ to learn a generative color prior by focusing on synthesizing colors on top of the spatial feature $f$.
The reduced learning complexity of the generator results in an enlarged representation space, covering in-the-wild natural images as demonstrated in Fig.~\ref{fig:ablation_fdim_teaser}(d).

\paragraph{Multi-modal Image Colorization}
Image colorization is an inherently ill-posed problem as  multiple potential color images could explain a single grayscale image.
We handle this multi-modal nature of image colorization by injecting the random code $z$ sampled from a normal distribution into the generator $G$.
Sampling multiple latent code $z$ enables synthesizing diverse color images.
Note that we do not provide the random code to the encoder as the multi-modal nature only applies to the color synthesis, not the spatial feature extraction.

\subsection{Training Details}
\subsubsection{Adversarial Training}
We train our framework in an alternating manner for adversarial learning.
We define our encoder-generator loss function $\mathcal{L}^{G}$ as a sum of three terms:
\begin{eqnarray}
    \mathcal{L}^{G} &=& \mathcal{L}_{mse}^{G} + \lambda_{per}\mathcal{L}_{per}^{G} + \lambda_{adv}\mathcal{L}_{adv}^{\mathcal{G}} ,
    \label{eq:Full}
\end{eqnarray}
where $\mathcal{L}_{mse}^{G}$ and $\mathcal{L}_{per}^{G}$ are the MSE reconstruction losses that penalize the color and perceptual discrepancies between the synthesized image $\hat{x}_{rgb}$ and the ground truth image ${x}_{rgb}$.
For the perceptual loss $\mathcal{L}_{per}^{G}$, we use the VGG16~\cite{VGG} features at 1st, 2nd, 6th, and 9th layers. 
$\mathcal{L}_{adv}^{\mathcal{G}}$ is the adversarial loss, specifically the class-conditional hinge loss~\cite{HingeLoss} defined as $\mathcal{L}_{adv}^{\mathcal{G}} = -D(\hat{x}_{rgb}, c)$.
We use the balancing weights $\lambda_{per}$ and $\lambda_{adv}$ set as 0.2 and 0.03 respectively.
For discriminator training, we also use the hinge loss~\cite{HingeLoss}
\begin{equation}
\label{eq:loss_d}
\mathcal{L}_{adv}^{\mathcal{D}} = -\mathrm{min}(0, -1+D({x}_{rgb}, c)) + \mathrm{min}(0, -1-D(\hat{x}_{rgb}, c)).
\end{equation}

\subsubsection{Color Augmentation}
\label{sec:Color Enhancement Augmentation}
To promote synthesizing vivid color, we apply a simple color augmentation to the real color images fed to the discriminator.
Specifically, we scale chromaticity of images in \textit{YUV} color space as $\{U, V\} \leftarrow \{1.2 U, 1.2 V\}$.
This color augmentation makes colors of semantically different regions in training images more distinguishable.
As a result, it helps the generator learn to synthesize not only more vivid but also semantically more correct colors,
which is not achievable by direct augmentation of generator output as will be shown in Sec.~\ref{sec:ablation}.
\section{Experiments}\label{sec:experiments}


\paragraph{Implementation}
We train our model on 1.2M color images of the ImageNet 1K~\cite{ImageNet} training set after excluding 10\% original images with low colorfulness scores~\cite{Colorfulness}.
We generate grayscale images based on a conventional linear-combination method\footnote{$L = 0.2989R + 0.5870G + 0.1140B$, where $L$ is the grayscale intensity and $R, G, B$ are the trichromatic color intensities.}.
We resize and crop the training images to be $256\times256$. 
For training, we use the Adam optimizer~\cite{Adam} with the coefficients of $\beta_1=0.0$ and $\beta_2=0.999$. 
The learning rates are set to $0.0001$ for the encoder-generator module and $0.00003$ for the discriminator with the decay rate of $0.9$ per epoch. We also use the exponential moving average~\cite{PGGAN} with the coefficient of $\beta = 0.999$ for model parameter update.
We set the batch size to $60$ and train the entire model for 12 epochs. 


\subsection{Evaluation}

We evaluate the effectiveness of \MethodName~on the ImageNet-1K validation set of 50K images~\cite{ImageNet} that have complex spatial structures.
\vspace{-5mm}

\subsubsection{Comparison with Other Colorization Methods}
\begin{figure}
\centering
\includegraphics[width=12cm]{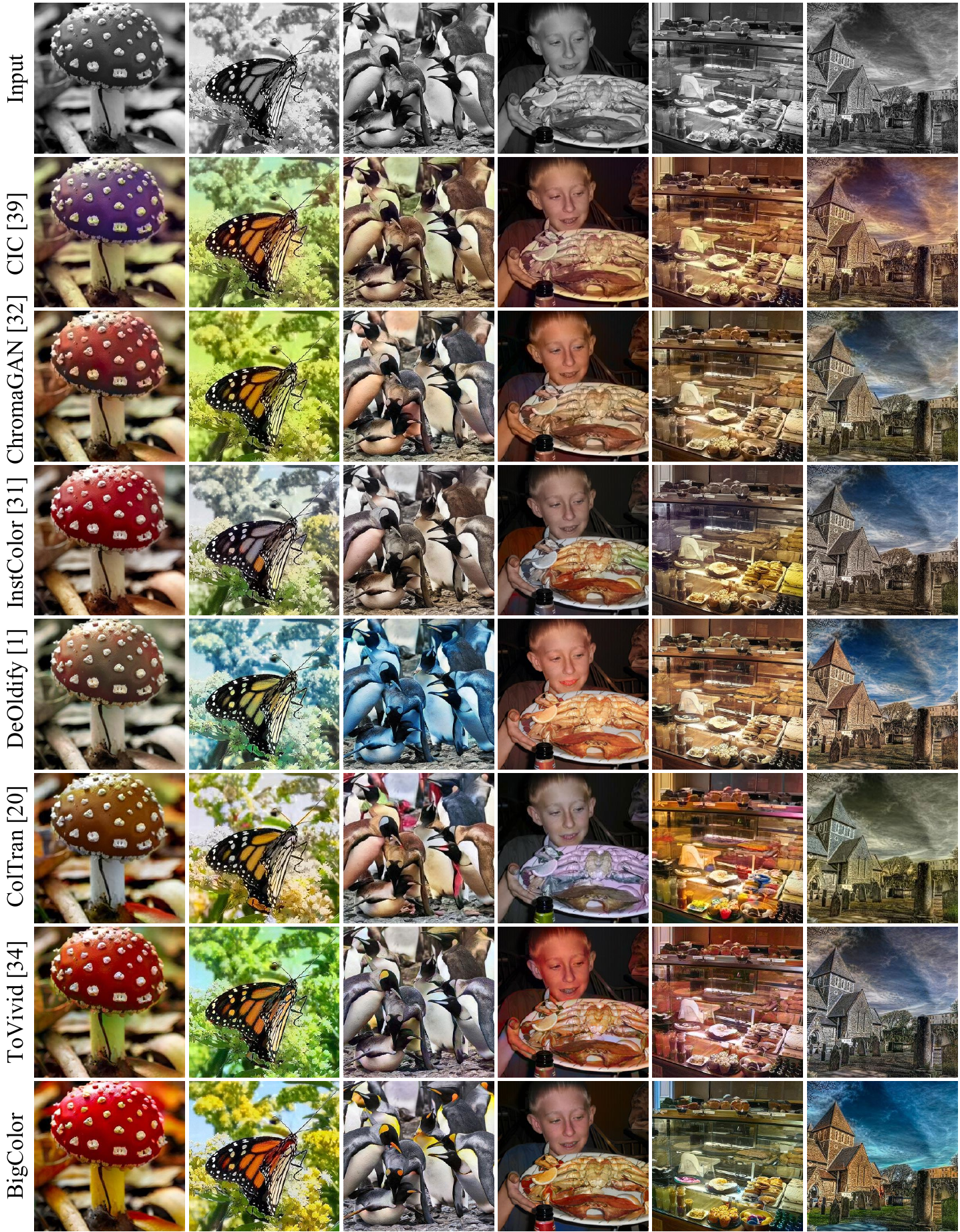}
\caption{Qualitative comparison with other colorization methods. For in-the-wild images with complex structure, our method synthesizes natural and vivid color images while the other methods suffer from desaturated and  unnatural color distributions. 
}
\label{fig:cmp}
\vspace{-3mm}
\end{figure}

We compare \MethodName\ to recent automatic colorization methods including
CIC~\cite{CIC}, ChromaGAN~\cite{ChromaGAN}, DeOldify~\cite{Deoldify}, InstColor~\cite{InstColor}, ColTran~\cite{ColTran} and ToVivid~\cite{TowardVivid}.
\Fig{\ref{fig:cmp}} shows that \MethodName\ qualitatively outperforms all the methods on six challenging images.
\MethodName\ successfully colorizes the complex structures of human faces, penguin heads, food, and buildings with semantically-natural and vivid colors.  
The two notable state-of-the-art methods of ToVivid~\cite{TowardVivid} and ColTran~\cite{ColTran} suffer from unnatural colorization as shown on the penguins and the human face due to their limited representation space.
This clearly demonstrates the effectiveness of our learned generative color prior to in-the-wild images. 
See the Supplemental Document for more qualitative results.

\begin{table}[t]
\begin{center}
\caption{Quantitative comparison with other colorization methods using the three metrics of colorfulness~\cite{Colorfulness}, FID~\cite{FID}, and classification accuracy~\cite{CIC}. \MethodName\ outperforms all previous work with significant margins. \textit{Aug.} denotes our color-augmentation scheme. The bold and underlined scores are the best and 2nd best results. 
}
\setlength{\tabcolsep}{1.em}
\begin{tabular}{c|*{4}{c}}

\Xhline{2\arrayrulewidth}
     & Colorful $\uparrow$ \cite{Colorfulness} & FID $\downarrow$ \cite{FID} & Classification $\uparrow$ \\
\hline\hline
CIC \cite{CIC}               & 33.036 & 11.322 & 69.976 \\
ChromaGAN \cite{ChromaGAN}   & 26.266 & 8.209 & 70.374  \\
InstColor \cite{InstColor}   & 25.507 & 7.890 & 68.422  \\
DeOldify \cite{Deoldify}     & 23.793 & 3.487 & 72.364 \\
ColTran \cite{ColTran}       & 34.485 & 3.793 & 67.210 \\
ToVivid \cite{TowardVivid}   & 35.128 & 4.078 & 73.816 \\
\hline
BigColor (w/o Aug.)              & \underline{36.157} & \underline{1.288} & \underline{76.302} \\
BigColor (w/ Aug.)               & {\bf 40.006} & {\bf 1.243} & {\bf 76.516}  \\
\Xhline{2\arrayrulewidth}

\end{tabular}
\label{table:cmp_metric}
\end{center}
\vspace{-5mm}
\end{table}
We further evaluate \MethodName\ using the three quantitative metrics of colorfulness, FID, and classification accuracy commonly used in the image colorization field.
Colorfulness measures the overall colorfulness of an image based on psychological experiments~\cite{Colorfulness}.
FID describes the distributional distance between the real color images and synthesized color images~\cite{FID}. 
The classification accuracy measures whether a classifier trained on natural color images, specifically the pretrained ResNet50-based classifier~\cite{ResNet}, can predict the correct classes of synthesized color images which were used in CIC~\cite{CIC}.
\Tbl{\ref{table:cmp_metric}} shows that \MethodName\ outperforms the previous methods with significant margins across all tested metrics with and without the color-augmentation scheme.

\paragraph{In-the-wild Images with Complex Structures}
We test the robustness of \MethodName\ specifically on challenging in-the-wild images with complex structures. 
To this end, we select 100 challenging images selected from the ImageNet1K validation set which contain as many humans as possible using an off-the-shelf object detector~\cite{Detectron2}, assuming the proportionality between the number of people and the image complexity.
On the curated dataset with 100 samples, \Tbl{\ref{table:cplximg}} shows the classification accuracy of the synthesized color images for all the methods. 
\MethodName\ again achieves the best performance with only a 2.5\% accuracy drop from the whole-data evaluation. 
Our performance drop of 2.5\% is at the same level of the ground-truth case, where real color images are used to obtain the classification accuracy. 
We refer to the Supplemental Document for further quantitative and qualitative evaluations. 

\begin{table}[t]
\begin{center}
\caption{\label{table:cplximg}
\MethodName\ is robust for colorizing complex images compared to the previous colorization methods, achieving the best performance in terms of classification accuracy with a marginal performance drop similar to the real ground-truth color images. 
}
\setlength{\tabcolsep}{1.5em}
\begin{tabular}{c|*{3}{c}}
\Xhline{2\arrayrulewidth}
 & \multicolumn{2}{c}{\textit{Classification acc.}}  & \textit{Performance drop}\\
Dataset & Whole & Complex & Whole $\rightarrow$ Complex \\
\hline\hline
Ground Truth                       & 78.530 & 76.000 & -2.530 \\
\hline
CIC~\cite{CIC}                     & 69.976 & 64.000 & -5.976 \\
ChromaGAN~\cite{ChromaGAN}         & 70.374 & 60.000 & -10.374 \\
InstColor~\cite{InstColor}         & 68.422 & 65.000 & -3.422 \\
DeOldify~\cite{Deoldify}           & 72.364 & 68.000 & -4.364 \\
ColTran~\cite{ColTran}             & 67.210 & 65.000 & -2.210 \\
ToVivid~\cite{TowardVivid}         & 73.816 & 65.000 & -8.816 \\
\hline
BigColor                        & {\bf 76.516} & {\bf 74.000} & -2.516 \\
\Xhline{2\arrayrulewidth}
\end{tabular}
\end{center}
\vspace{-3mm}
\end{table}

\begin{figure}[t]
  \begin{center}
    \includegraphics[width=11.5cm]{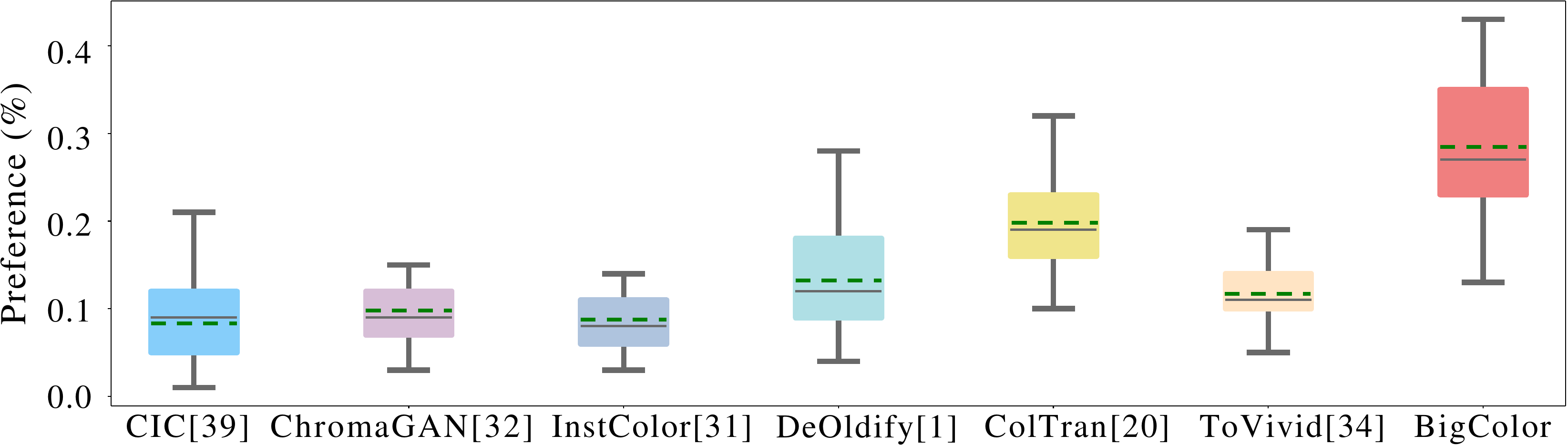}
    \vspace{-3mm}
    \end{center}
    \caption{
    We conduct a user study to evaluate the preference for colorization results. 
    In all tested metrics, our method outperforms the other methods. 
    The dashed green line and the bold gray line inside the bars are the mean and the median respectively.
    }
    \label{fig:userstudy}
\end{figure}

\def\NumParticipant{33}
\def\NumImg{100}
\paragraph{User Study}
We conducted a user study to investigate the perceptual preference of colorization methods using Amazon Mechanical Turk (AMT).
Specifically, {\NumParticipant} subjects are presented with {\NumImg} input and colorized images randomly selected from the ImageNet 1K validation set.
The subjects choose the best-restored color image among the results obtained with different methods~\cite{CIC,ChromaGAN,InstColor,Deoldify,ColTran,TowardVivid}.
\Fig{\ref{fig:userstudy}} shows that users clearly prefer \MethodName\ over the state-of-the-art methods. 
More details can be found in the Supplemental Document.


\subsection{Ablation Study}
\label{sec:ablation}
We conduct extensive ablation studies to assess \MethodName\ in details by using 10\% of the ImageNet training images amounting to 100 image classes. 
\vspace{-5mm}


\begin{figure}[t]
\centering
\includegraphics[width=12cm]{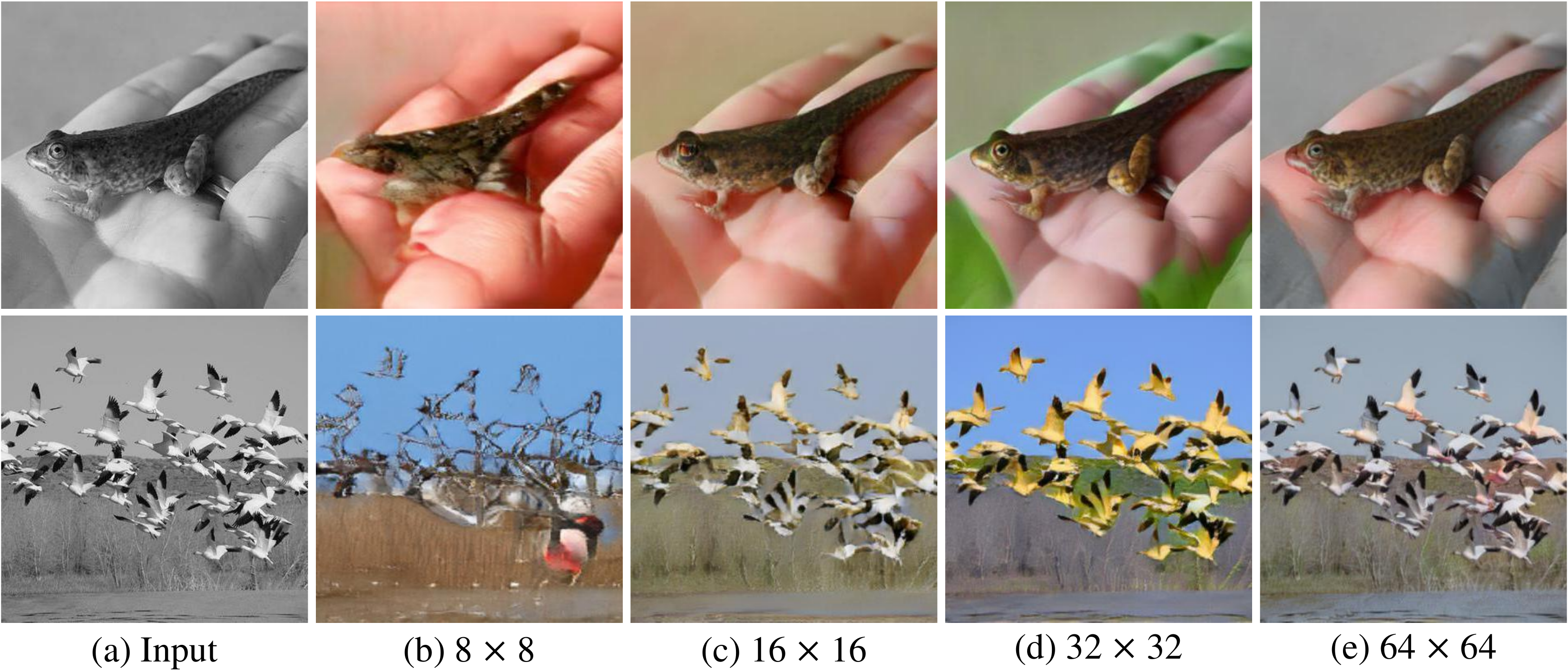}
\caption{
The resolution of the spatial feature $f$ plays an important role for maintaining a balance between keeping the spatial structure of the input and providing the degree of freedom for synthesizing color. We empirically chose $16 \times 16$ as the best configuration. 
}
\vspace{-2mm}
\label{fig:ablation_fdim}
\end{figure}
\subsubsection{Resolution of the Spatial Feature}
\label{sec:FeatureSpaceResolutions}
We evaluate the impact of the resolution of the spatial feature $f$. 
\Fig{\ref{fig:ablation_fdim}} shows the colorization results with varying spatial resolutions of the feature $f$ from $8\times8$ to $64 \times 64$.
As the spatial resolution increases, the synthesized color images can exploit more structural information of the input image for colorization.
However, a large spatial resolution could harm the colorization results as it reduces the capacity of the generator with fewer layers. 
We chose $16\times16$ as the spatial resolution of the feature $f$.  


\begin{figure}[t]
\centering
\includegraphics[width=12cm]{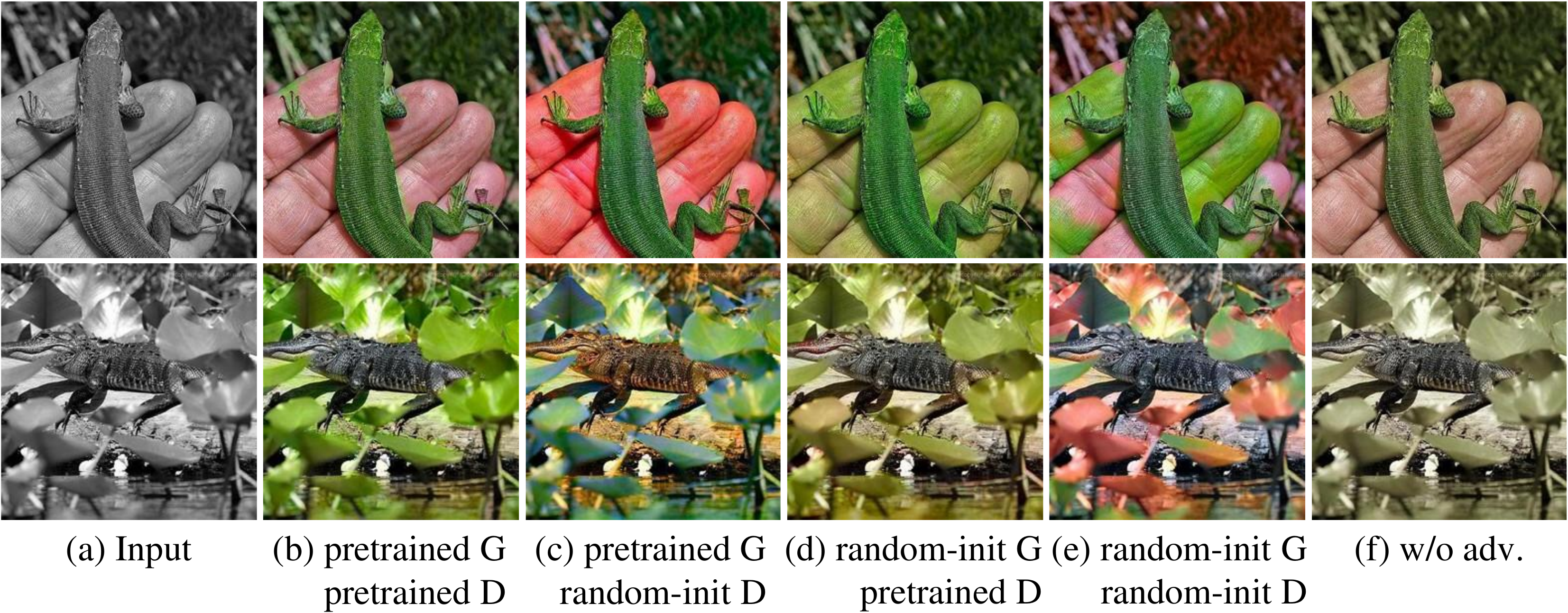}
\vspace{-2mm}
\caption{
We evaluate the impact of initializing the generator $G$ and the discriminator $D$ with the pretrained BigGAN model.
Compared to the random initialization, pretrained initialization results in vivid and natural colorization. Also, adversarial training is critical to achieving vivid colorization without desaturation. 
}
\label{fig:ablation_prior}
\end{figure}
\begin{table}[t]
\begin{center}
\caption{
Initialization with the pretrained BigGAN model provides quantiatively better results in terms of FID and classification accuracy. 
}
\vspace{-2mm}
\setlength{\tabcolsep}{0.9em}
\begin{tabular}{c|cc|cc|cc|cc|c}
\Xhline{2\arrayrulewidth}
 Model & G & D & G & D & G & D & G & D &\multirow{2}{*}{w/o adv.} \\\cline{1-9}
 Pretrained & \checkmark & \checkmark & \checkmark & - & - & \checkmark & - & - &  \\\cline{2-9}
\hline\hline
FID~\cite{FID}   & \multicolumn{2}{c}{\bf 5.714} & \multicolumn{2}{c}{8.058} & \multicolumn{2}{c}{6.852} & \multicolumn{2}{c}{7.201} & \multicolumn{1}{c}{7.692} \\
Class. Acc.              & \multicolumn{2}{c}{\bf 81.44} & \multicolumn{2}{c}{78.96} & \multicolumn{2}{c}{80.78} & \multicolumn{2}{c}{80.60} & \multicolumn{1}{c}{75.52} \\
\Xhline{2\arrayrulewidth}
\end{tabular}
\label{table:ablation_prior}
\end{center}
\vspace{-7mm}
\end{table}

\begin{table}[t]
\begin{center}
\caption{We analyze our encoder architecture in details to provide insight on the importance of each encoder component: batch normalization (BN), class-conditioned batch normalization (CBN), residual learning (RL). 
The encoder with residual path and class-conditioned batch normalization shows the best result in terms of FID.
}
\vspace{-1mm}
\setlength{\tabcolsep}{1.5em}
\begin{tabular}{c||*{3}{c}}
\Xhline{2\arrayrulewidth}
 & w/o BN & w/ BN & w/ CBN \\
\hline\hline
w/o RL.    & 6.523 & 7.286 & 5.974 \\
w/ RL.     & 5.980 & 5.854 & {\bf 5.714}  \\
\Xhline{2\arrayrulewidth}
\end{tabular}
\label{table:ablation_encoder}
\end{center}
\vspace{-3mm}
\end{table}

\paragraph{Initialization with a Pretrained Generative Prior}
\label{sec:GenerativePrior}
We initialize our generator and discriminator using the BigGAN pretrained model in order to leverage the learned structure-color distribution of natural images.
\Fig{\ref{fig:ablation_prior}} and \Tbl{\ref{table:ablation_prior}} show that the pretrained initialization improves performance over the training-from-scratch alternatives with random initialization. 
Specifically, we test all four combinations of the generator-discriminator initialization settings with and without the pretrained initialization.
The qualitative and quantitative results indicate that \MethodName~successfully exploits the pretrained information in the BigGAN generator and the discriminator.
We also confirmed the importance of including the adversarial loss to achieve vivid colorization.

\paragraph{Encoder Architecture}
\label{sec:EncoderArchitecture} 
We considered two main factors for designing our encoder architecture: extracting image structure and exploiting class information.
We found that the residual blocks and class-conditioned batch normalization in the original BigGAN generator are essential for robust image colorization as shown in \Tbl{\ref{table:ablation_encoder}}.
Specifically, residual blocks transfer structural information and the class-conditioned batch normalization extracts the class-specific spatial feature. 

\begin{table}[t]
\begin{center}
\caption{\label{table:coloraug}
Augmenting the real color image fed to the discriminator improves the colorization performance measured in FID and classification accuracy. Our experiments also confirm that directly augmenting the synthesized color degrades the colorization performance.
Disc. and Gen. denote the color augmentation on the real color image fed to the discriminator and the generated color image respectively.
}
\vspace{-2mm}
\setlength{\tabcolsep}{0.7em}
\begin{tabular}{c|cc|cc|cc|cc}
\Xhline{2\arrayrulewidth}
 \multirow{2}{*}{Color Aug.} & Disc. & Gen. & Disc. & Gen. & Disc. & Gen. & Disc. & Gen.\\\cline{2-9}
  & \checkmark & - & \checkmark & \checkmark & - & - & - & \checkmark \\\cline{2-8}
\hline\hline
FID          & \multicolumn{2}{c}{\bf 1.243} & \multicolumn{2}{c}{1.604} & \multicolumn{2}{c}{1.288} & \multicolumn{2}{c}{1.621}  \\
Class. Acc.  & \multicolumn{2}{c}{\bf 76.516} & \multicolumn{2}{c}{76.282} & \multicolumn{2}{c}{76.302} & \multicolumn{2}{c}{76.238} \\
\Xhline{2\arrayrulewidth}
\end{tabular}
\end{center}
\vspace{-7mm}
\end{table}
\begin{figure}[t]
\centering
\includegraphics[width=12cm]{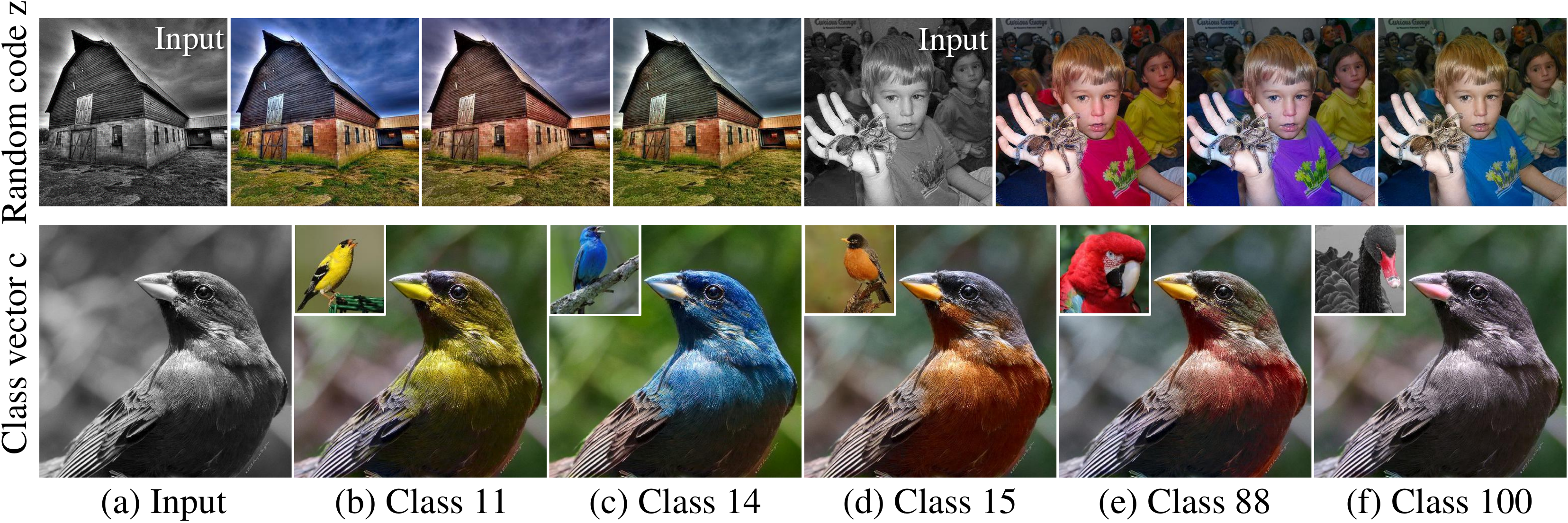}
\vspace{-2mm}
\caption{
\MethodName\ supports multi-modal image colorization by sampling the random code $z$ or using different class vectors $c$ which can be estimated from the reference images shown in the insets. The class indices estimated from the reference images are shown below each of the colorization results.
}
\vspace{-2mm}
\label{fig:diverse_color}
\end{figure}
\begin{figure}[t]
\centering
\includegraphics[width=12cm]{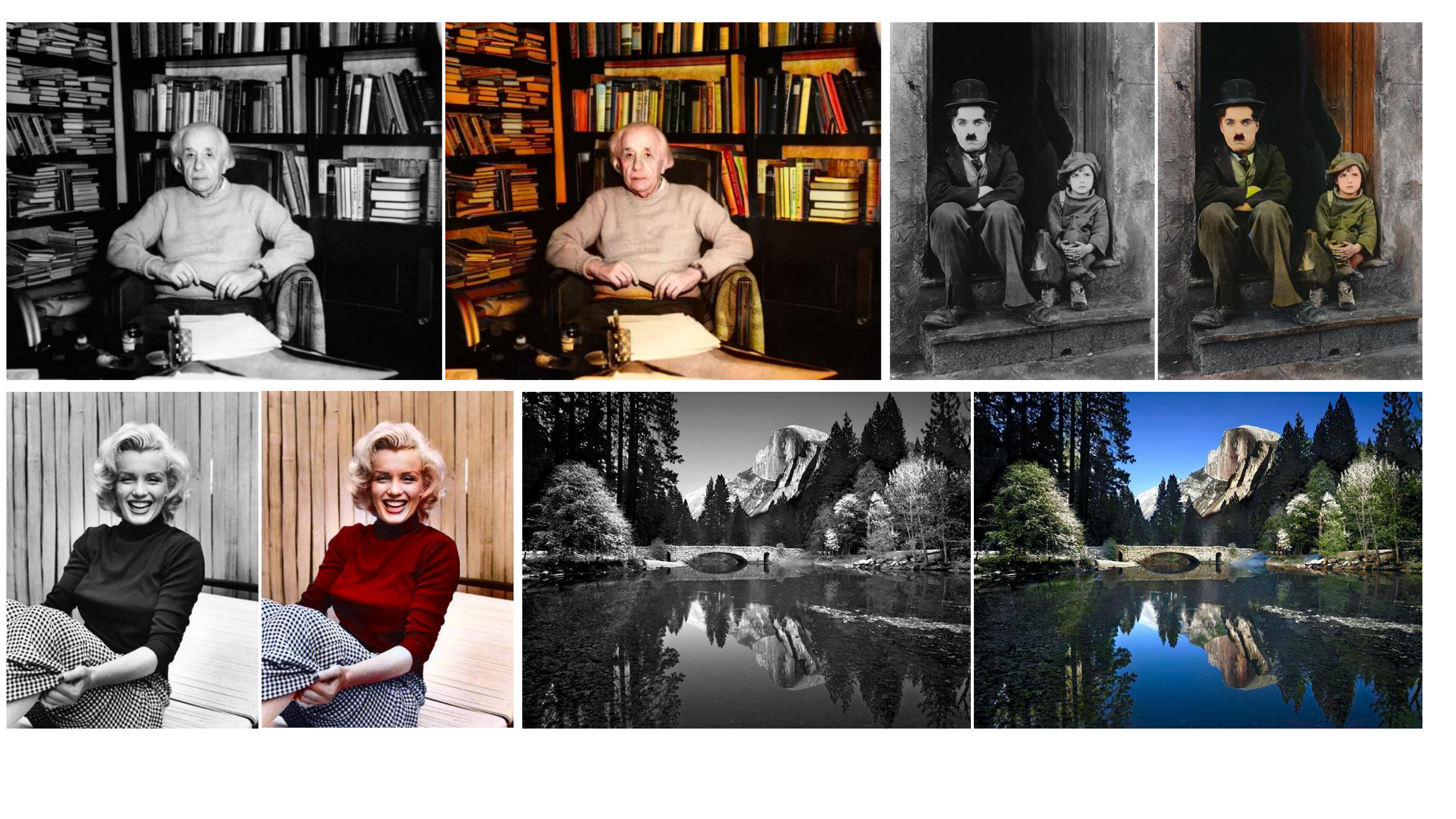}
\vspace{-10mm}
\caption{We apply \MethodName\ to old monochromatic photographs of diverse resolutions and aspect ratios. Left-top to right-bottom: {\it Albert Einstein at Princeton University, Charlie Chaplin on the movie `The Kid'(1921), Marilyn Monroe, Photo by Ansel Adams of Yosemite.}}
\label{fig:real}
\vspace{-3mm}
\end{figure}

\if{0}
\caption{\MethodName\ on old monochromatic photographs. Left-top to right-bottom: {\it Charlie Chaplin on movie `The Kid'(1921), Migrant Mother by Dorothea Lange(1936), Albert Einstein at Princeton University, Mrilyn Monroe}}
\fi
\paragraph{Color Augmentation}
We experimentally evaluate the impact of color augmentation on the real color images fed to the discriminator.
To this end, we compare the FID score and the classification accuracy on 1000 classes of the ImageNet with and without the color augmentation, which shows clear improvements in both metrics as shown in \Tbl{\ref{table:coloraug}}. 
We also test applying the color augmentation on the synthesized image from the generator as a post-processing after training.
This does not consider image semantics, resulting in unnatural colorization as indicated by the FID and the classification scores. 
In contrast, augmenting the discriminator input enables us to effectively learn the vivid and semantically correct color distribution of the real images.
More discussion with qualitative examples of the color augmentation is provided in the Supplemental Document.



\subsection{Multi-modal Colorization}
\label{sec:ColorEditing}
\MethodName\ is capable of synthesizing diverse colorization results for an input grayscale image as shown in \Fig{\ref{fig:diverse_color}}. 
We can sample random code $z$ that is injected into the generator to synthesize diverse color images. 
In addition, we can also alter the class vector $c$ to generate class-specific colorization results, for instance by using the class codes of different classes of birds to colorize an input bird image as shown in the second row in \Fig{\ref{fig:diverse_color}}.

\vspace{-3mm}
\subsection{Black-and-White Photo Restoration}
 \Fig{\ref{fig:real}} shows the colorization results of \MethodName\ for old monochromatic photographs with arbitrary resolutions and aspect ratios.
Note that \MethodName\ is not limited to a specific input resolution owing to the convolutional spatial feature $f$ with a variable spatial resolution.
In contrast, conventional GAN-inversion methods~\cite{TowardVivid,DGP} use a spatially-flattend latent code, enforcing the spatial resolution to be fixed.  

\vspace{-3mm}
\section{Conclusion}
\vspace{-2mm}
\label{sec:conclusion}
We propose \MethodName, a robust image colorization method using a generative color prior for in-the-wild images with complex structures.
We exploit the spatial structure of an input grayscale image using a convolutional encoder, effectively enlarging the representation space of a generator compared to the conventional colorization methods using GAN inversion.
Jointly optimizing the encoder-generator module with a discriminator allows us to learn a generative color prior where the generator focuses on synthesizing colors on top of the extracted spatial-structure feature. 
We extensively assess \MethodName\ in qualitative and quantitative manners and demonstrate that \MethodName\ outperforms existing state-of-the-art methods. 


\vspace{-0mm}
\paragraph{Limitations} 
Our method is not free from limitations.
The spatial resolution of the extracted feature $f$ determines the structural details that can be maintained for the color synthesis procedure. 
Thus, tiny regions might be overlooked in the colorization process.  Also, we rely on the BigGAN class code which may not be perfectly estimated for challenging images. 

\paragraph{\small Acknowledgements}
This work was supported by the National Research Foundation of Korea (NRF) grant funded by the Korea government (MSIT) (NRF-2018R1A5A1060031), Institute of Information \& communications Technology Planning \& Evaluation (IITP) grant funded by the Korea government (MSIT) (No.2019-0-01906, Artificial Intelligence Graduate School Program(POSTECH)), and Samsung Electronics Co., Ltd.


\bibliographystyle{splncs04}
\bibliography{main.bbl}

\begin{thebibliography}{10}
\providecommand{\url}[1]{\texttt{#1}}
\providecommand{\urlprefix}{URL }
\providecommand{\doi}[1]{https://doi.org/#1}

\bibitem{Deoldify}
Antic, J.: Deoldify. \url{https://github.com/jantic/DeOldify} (2019)

\bibitem{BigGAN}
Brock, A., Donahue, J., Simonyan, K.: Large scale gan training for high
  fidelity natural image synthesis. In: International Conference on Learning
  Representations (2019)

\bibitem{GLEAN}
Chan, K.C., Wang, X., Xu, X., Gu, J., Loy, C.C.: Glean: Generative latent bank
  for large-factor image super-resolution. In: Proceedings of the IEEE/CVF
  Conference on Computer Vision and Pattern Recognition. pp. 14245--14254
  (2021)

\bibitem{charpiat2008automatic}
Charpiat, G., Hofmann, M., Sch{\"o}lkopf, B.: Automatic image colorization via
  multimodal predictions. In: European conference on computer vision. pp.
  126--139. Springer (2008)

\bibitem{DeepColor}
Cheng, Z., Yang, Q., Sheng, B.: Deep colorization. In: Proceedings of the IEEE
  international conference on computer vision. pp. 415--423 (2015)

\bibitem{chia2011semantic}
Chia, A.Y.S., Zhuo, S., Gupta, R.K., Tai, Y.W., Cho, S.Y., Tan, P., Lin, S.:
  Semantic colorization with internet images. ACM Transactions on Graphics
  (TOG)  \textbf{30}(6), ~1--8 (2011)

\bibitem{LargeAutomaticColor}
Deshpande, A., Rock, J., Forsyth, D.: Learning large-scale automatic image
  colorization. In: Proceedings of the IEEE international conference on
  computer vision. pp. 567--575 (2015)

\bibitem{mGANPrior}
Gu, J., Shen, Y., Zhou, B.: Image processing using multi-code gan prior. In:
  Proceedings of the IEEE/CVF conference on computer vision and pattern
  recognition. pp. 3012--3021 (2020)

\bibitem{gupta2012image}
Gupta, R.K., Chia, A.Y.S., Rajan, D., Ng, E.S., Zhiyong, H.: Image colorization
  using similar images. In: Proceedings of the 20th ACM international
  conference on Multimedia. pp. 369--378 (2012)

\bibitem{Colorfulness}
Hasler, D., Suesstrunk, S.E.: Measuring colorfulness in natural images. In:
  Human vision and electronic imaging VIII. vol.~5007, pp. 87--95.
  International Society for Optics and Photonics (2003)

\bibitem{ResNet}
He, K., Zhang, X., Ren, S., Sun, J.: Deep residual learning for image
  recognition. In: Proceedings of the IEEE conference on computer vision and
  pattern recognition. pp. 770--778 (2016)

\bibitem{FID}
Heusel, M., Ramsauer, H., Unterthiner, T., Nessler, B., Hochreiter, S.: Gans
  trained by a two time-scale update rule converge to a local nash equilibrium.
  Advances in neural information processing systems  \textbf{30} (2017)

\bibitem{huang2005adaptive}
Huang, Y.C., Tung, Y.S., Chen, J.C., Wang, S.W., Wu, J.L.: An adaptive edge
  detection based colorization algorithm and its applications. In: Proceedings
  of the 13th annual ACM international conference on Multimedia. pp. 351--354
  (2005)

\bibitem{LetThereBeColor}
Iizuka, S., Simo-Serra, E., Ishikawa, H.: Let there be color! joint end-to-end
  learning of global and local image priors for automatic image colorization
  with simultaneous classification. ACM Transactions on Graphics (ToG)
  \textbf{35}(4),  1--11 (2016)

\bibitem{BDInvert}
Kang, K., Kim, S., Cho, S.: Gan inversion for out-of-range images with
  geometric transformations. In: Proceedings of the IEEE/CVF International
  Conference on Computer Vision. pp. 13941--13949 (2021)

\bibitem{PGGAN}
Karras, T., Aila, T., Laine, S., Lehtinen, J.: Progressive growing of gans for
  improved quality, stability, and variation. arXiv preprint arXiv:1710.10196
  (2017)

\bibitem{StyleGAN}
Karras, T., Laine, S., Aila, T.: A style-based generator architecture for
  generative adversarial networks. In: Proceedings of the IEEE/CVF conference
  on computer vision and pattern recognition. pp. 4401--4410 (2019)

\bibitem{StyleGAN2}
Karras, T., Laine, S., Aittala, M., Hellsten, J., Lehtinen, J., Aila, T.:
  Analyzing and improving the image quality of stylegan. In: Proceedings of the
  IEEE/CVF Conference on Computer Vision and Pattern Recognition. pp.
  8110--8119 (2020)

\bibitem{Adam}
Kingma, D.P., Ba, J.: Adam: A method for stochastic optimization. arXiv
  preprint arXiv:1412.6980  (2014)

\bibitem{ColTran}
Kumar, M., Weissenborn, D., Kalchbrenner, N.: Colorization transformer. In:
  International Conference on Learning Representations (2021)

\bibitem{LearningAumotaticColor}
Larsson, G., Maire, M., Shakhnarovich, G.: Learning representations for
  automatic colorization. In: European conference on computer vision. pp.
  577--593. Springer (2016)

\bibitem{levin2004colorization}
Levin, A., Lischinski, D., Weiss, Y.: Colorization using optimization. In: ACM
  SIGGRAPH 2004 Papers, pp. 689--694 (2004)

\bibitem{HingeLoss}
Lim, J.H., Ye, J.C.: Geometric gan. arXiv preprint arXiv:1705.02894  (2017)

\bibitem{liu2008intrinsic}
Liu, X., Wan, L., Qu, Y., Wong, T.T., Lin, S., Leung, C.S., Heng, P.A.:
  Intrinsic colorization. In: ACM SIGGRAPH Asia 2008 papers, pp.~1--9 (2008)

\bibitem{Rephotography}
Luo, X., Zhang, X., Yoo, P., Martin-Brualla, R., Lawrence, J., Seitz, S.M.:
  Time-travel rephotography. ACM Transactions on Graphics (Proceedings of ACM
  SIGGRAPH Asia 2021)  \textbf{40}(6) (12 2021)

\bibitem{PULSE}
Menon, S., Damian, A., Hu, S., Ravi, N., Rudin, C.: Pulse: Self-supervised
  photo upsampling via latent space exploration of generative models. In:
  Proceedings of the ieee/cvf conference on computer vision and pattern
  recognition. pp. 2437--2445 (2020)

\bibitem{DGP}
Pan, X., Zhan, X., Dai, B., Lin, D., Loy, C.C., Luo, P.: Exploiting deep
  generative prior for versatile image restoration and manipulation. IEEE
  Transactions on Pattern Analysis and Machine Intelligence  (2021)

\bibitem{qu2006manga}
Qu, Y., Wong, T.T., Heng, P.A.: Manga colorization. ACM Transactions on
  Graphics (TOG)  \textbf{25}(3),  1214--1220 (2006)

\bibitem{ImageNet}
Russakovsky, O., Deng, J., Su, H., Krause, J., Satheesh, S., Ma, S., Huang, Z.,
  Karpathy, A., Khosla, A., Bernstein, M., et~al.: Imagenet large scale visual
  recognition challenge. International journal of computer vision
  \textbf{115}(3),  211--252 (2015)

\bibitem{VGG}
Simonyan, K., Zisserman, A.: Very deep convolutional networks for large-scale
  image recognition. arXiv preprint arXiv:1409.1556  (2014)

\bibitem{InstColor}
Su, J.W., Chu, H.K., Huang, J.B.: Instance-aware image colorization. In:
  Proceedings of the IEEE/CVF Conference on Computer Vision and Pattern
  Recognition. pp. 7968--7977 (2020)

\bibitem{ChromaGAN}
Vitoria, P., Raad, L., Ballester, C.: Chromagan: Adversarial picture
  colorization with semantic class distribution. In: Proceedings of the
  IEEE/CVF Winter Conference on Applications of Computer Vision. pp. 2445--2454
  (2020)

\bibitem{GFP-GAN}
Wang, X., Li, Y., Zhang, H., Shan, Y.: Towards real-world blind face
  restoration with generative facial prior. In: Proceedings of the IEEE/CVF
  Conference on Computer Vision and Pattern Recognition. pp. 9168--9178 (2021)

\bibitem{TowardVivid}
Wu, Y., Wang, X., Li, Y., Zhang, H., Zhao, X., Shan, Y.: Towards vivid and
  diverse image colorization with generative color prior. In: Proceedings of
  the IEEE/CVF International Conference on Computer Vision. pp. 14377--14386
  (2021)

\bibitem{Detectron2}
Wu, Y., Kirillov, A., Massa, F., Lo, W.Y., Girshick, R.: Detectron2.
  \url{https://github.com/facebookresearch/detectron2} (2019)

\bibitem{xu2009efficient}
Xu, K., Li, Y., Ju, T., Hu, S.M., Liu, T.Q.: Efficient affinity-based edit
  propagation using kd tree. ACM Transactions on Graphics (TOG)
  \textbf{28}(5), ~1--6 (2009)

\bibitem{GPEN}
Yang, T., Ren, P., Xie, X., Zhang, L.: Gan prior embedded network for blind
  face restoration in the wild. In: Proceedings of the IEEE/CVF Conference on
  Computer Vision and Pattern Recognition. pp. 672--681 (2021)

\bibitem{yatziv2006fast}
Yatziv, L., Sapiro, G.: Fast image and video colorization using chrominance
  blending. IEEE transactions on image processing  \textbf{15}(5),  1120--1129
  (2006)

\bibitem{CIC}
Zhang, R., Isola, P., Efros, A.A.: Colorful image colorization. In: European
  conference on computer vision. pp. 649--666. Springer (2016)

\end{thebibliography}

\end{document}